# Business World Model

Cecil Pang[1,2] and Hiroki Sayama[2,3,4]

[1]AUTOBUS Lab, Inc., New York, USA
[2]School of Systems Science and Industrial Engineering, Binghamton University, State University of New York, USA
[3]Binghamton Center of Complex Systems, Binghamton University, State University of New York, USA
[4]Waseda Innovation Lab, Waseda University, Tokyo, Japan

ORCID (Cecil Pang): 0009-0003-1725-7873 ORCID (Hiroki Sayama): 0000-0002-2670-5864

**Abstract**

World model has emerged as a powerful paradigm in artificial intelligence, enabling agents to represent their environments, predict future states, and evaluate possible actions before acting. However, existing world models have largely been developed for visual or physical domains such as computer vision, robotics, gaming, and autonomous driving, which are governed by relatively stable dynamics. These formulations are not directly applicable to business practice, where the relevant environment is semantic, organizational, and market driven. Outcomes depend on context-sensitive factors such as customer behavior, pricing, competition, and regulation. This paper introduces the concept and architecture of a Business World Model, a world model specialized for business and organizational environments. We propose a business-semantics-centric formulation in which states, dynamics, and feasible actions are linked to business entities, their attributes, and their relationships. Within this framework, intelligent agents can simulate alternative action sequences, estimate their effects on future business outcomes, and evaluate trade-offs under uncertainty and business constraints. We present a reference design and open-source implementation in which LLM agents interpret natural-language business goals and explain results, while a deterministic, metadata-driven simulation engine integrating machine learning models, business rules, and action spaces performs all calculations. A churn case study of a digital subscription business demonstrates that the system recommends options meeting the churn-reduction target within the revenue constraint. This work establishes a conceptual foundation for autonomous business systems capable of moving from instruction-based execution toward goal-driven planning, optimization, and execution.

## 1 Introduction

### 1.1 Motivation

World models have emerged as an important paradigm in artificial intelligence for enabling agents to represent their environments, predict future states, and evaluate possible actions before acting. They have been widely studied in domains such as computer vision, robotics, gaming, autonomous driving, and embodied AI, where the "world" is often modeled as a visual or physical environment governed by the laws of physics. In such settings, a world model may learn how objects move, how scenes evolve, or how an agent's physical actions affect its surroundings. A planner then uses this world model to plan and optimize multi-step actions, or to train a policy that will be used to guide the actions (Ha and Schmidhuber 2018).

Existing world model formulations are not directly applicable to business practice. Unlike physical environments, business environments are not governed by universal and stable laws analogous to physics. Business outcomes depend on context-sensitive and often changing factors, including customer behavior, organizational constraints, market competition, pricing, regulation, incentives, and macroeconomic conditions. Moreover, the relevant "state" of a business is not primarily visual or spatial, it is semantic. It consists of entities such as customers, products, and subscriptions, along with their attributes and relationships, and dynamics among them. This creates the need for a specialized form of world model that can represent business states, reason over feasible business actions, and simulate their likely effects on future business outcomes.

Meanwhile, businesses have rapidly adopted AI-enabled tools to introduce new product capabilities, improve productivity, reduce operating costs, or pursue all these objectives simultaneously. We argue, however, that the significance of AI extends well beyond incremental tool support. Contemporary AI systems have the potential to ignite business growth by fundamentally changing how businesses are planned and executed. Rather than requiring executives to specify detailed instructions for every task within a business initiative, future business systems could operate from strategic outcome specifications, with intelligent agents performing planning and optimization to decompose goals into interdependent tasks, executing those tasks, and adapting as internal or external conditions change.

Consider a digital subscription company seeking to reduce customer churn. In today's instruction-based setting, a business initiative might be defined as follows:

*"Identify subscribers who pay more than $10 per month, are in churn-risk tier 4, and live in cities where their household incomes are above the median, then send them a 10% discount."*

According to a recent study (Pang and Sayama 2026a), a neuro-symbolic AI system can autonomously and reliably execute an initiative of this form. This is an important capability, but it remains execution-oriented: the plan, targeting logic, and campaign parameters are still specified in advance by the user. The system carries out the initiative, but it does not determine whether this is the best initiative to pursue, whether alternative targeting rules would be more effective, or how the campaign would affect other business outcomes such as revenue.

A world model developed for vision or robotics may learn to simulate how a physical scene changes in response to an action, such as how an object moves when pushed or how a vehicle's surroundings evolve as it turns. However, it is not designed to reason about business questions such as which subscribers should receive a discount, what discount level should be offered, how a marketing campaign would affect churn probability, or whether the resulting revenue impact would violate business constraints. These questions require a model of a different kind of "world": one composed not primarily of physical objects and spatial dynamics, but of business entities, semantic relationships, feasible interventions, uncertain behavioral responses, and operational constraints. To address this gap, we introduce Business World Model (BWM). The purpose of a BWM is to support planning, counterfactual reasoning, and the autonomous execution of business initiatives, not merely to support the execution of predefined instructions. It enables agents to reason about alternative courses of action before execution. Figure 1 shows that BWM plays the same functional role as conventional world models, but in a business context.

In a BWM-supported autonomous business system (AUTOBUS), observations received from the business environment are used to build and continuously update the BWM. Drawing on this representation, BWM predicts possible future states and evaluates the likely consequences of alternative actions before committing to one. BWM goes beyond simple reactions, it reasons over the modeled states and dynamics to plan multi-step business initiatives, simulate their outcomes, and select the most promising course of actions to execute. Once an action is taken, it changes the environment and leads to new observations and events. The new observations and events are fed back into the BWM, allowing the agent to adapt its understanding of the environment and refine future action plans. In this way, the BWM serves as the internal simulator that enables intelligent agents to autonomously plan, optimize, and execute business initiatives in pursuit of desired outcomes.

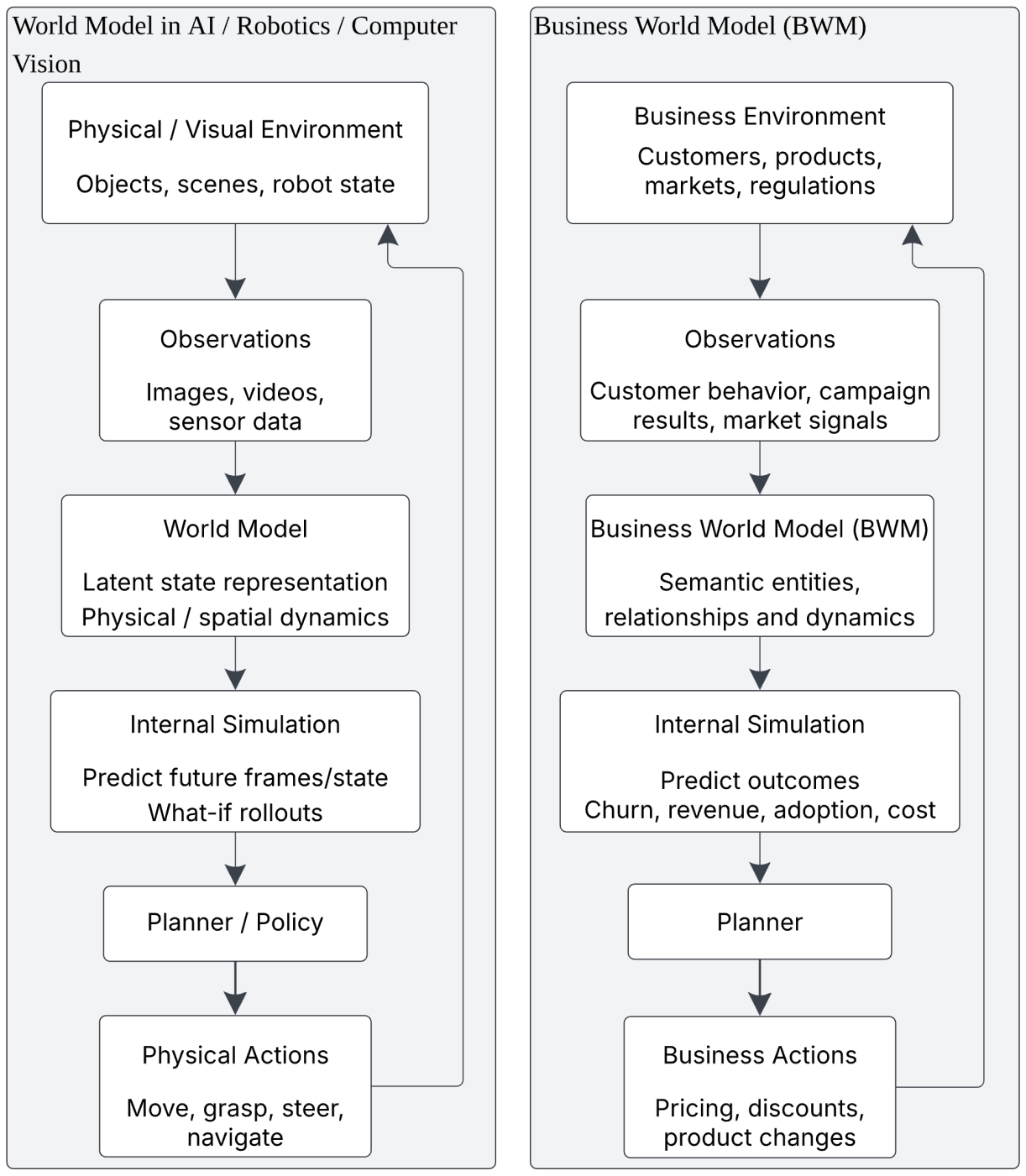

**Figure 1.** Functional analogy between conventional world model and Business World Model (BWM). Conventional world models in AI, robotics, and computer vision enable agents to represent physical or visual environments, simulate future states, and select actions before acting. A BWM serves a similar role for business environments, but its state representation is semantic rather than visual or physical. It models business entities, relationships, and dynamics so that intelligent agents can simulate alternative business initiatives, evaluate their likely outcomes and side effects, and plan actions toward strategic objectives.

With a BWM supporting the underlying planning and simulation mechanism, we propose moving beyond the autonomous execution of predefined plans. Instead of requiring human users to define exact execution logic and parameters, a BWM allows them to specify desired business outcomes:

*"Reduce the overall churn rate of paid subscribers by 2%. Propose two campaigns. The campaign may target no more than 20,000 subscribers, and total revenue must not decline by more than 0.01%."*

BWM would then generate candidate campaign options, estimate their likely effects, evaluate trade-offs, and execute the option selected by the user. For example, it might propose:

- *Candidate Campaign 1: Target subscribers who pay more than $10 per month, are in churn-risk tier 4, and have household income above their city's median. Offer them a 10% discount. The model estimates that this campaign would reduce the paid-subscriber churn rate by 2.1%, with no revenue loss.*
- *Candidate Campaign 2: Target subscribers who pay more than $15 per month, are in churn-risk tier 3, and have household income above their city's median. Offer them a 10% discount. The model estimates that this campaign would reduce the paid-subscriber churn rate by 2.1%, while decreasing revenue by 0.01%.*

To generate such candidate campaigns, BWM must answer questions such as what actions are feasible, which customer segments can be targeted, how would different discounts affect churn, and how would those actions affect revenue. It must also propose plans that satisfy the stated constraints. Answering these questions requires an internal model of the business environment that can support counterfactual reasoning and scenario simulation. This is the role of the BWM. It provides a structured representation of business states and dynamics and enables AUTOBUS to move from instruction-based execution toward goal-driven planning and optimization.

### 1.2 Contribution

This paper presents the concept and architecture of BWM as a foundational component of autonomous business system. The research problem is architectural and systemic because no single technology provides a solution. To the best of our knowledge, this work is the first to introduce a systematic Business World Model. We begin by reviewing world models in the literature and their historical foundations. Building on this background, we introduce the concept of a BWM and articulate the problem context. We then present the architecture of BWM in a modern data-rich business environment, followed by reference design and a case study.

## 2 Background

### 2.1 What is a World Model

A world model in the context of intelligent agents is an internal representation of the world that the agents use to reason about the environment (Ha and Schmidhuber 2018; Hafner et al. 2019). It supports understanding the current state as well as predicting future states of self and the world. In other words, it models the spatial and temporal dynamics of the environment: how states evolve over time, what plausible future states are, how actions influence them, and how observations relate to underlying states. When equipped with a world model, an agent can simulate possible futures, evaluate hypothetical actions, and select one that maximizes the likelihood of achieving a given goal. This leads to more intelligent behavior than simply reacting to current observations at every time step.

One main motivation for world models is inspired by biological intelligence**.** Humans and animals do not learn solely by going through all the scenarios by trial and error in the physical world, and do not need to know for certain all possible consequences before taking actions. This is especially true when there is perceived immediate danger. Rather, they mentally rehearse actions, perhaps unconsciously, imagine consequences based on the current understanding of the environment (which could be limited and full of uncertainties), generalize when necessary, make decisions, and then take actions. This is in essence a planning process. This also implies that some model of the world must exist in the mind so that carrying out this planning process is possible. When these world models are incomplete or uncertain, surprises may arise when predictions are different from what are observed from the environment (Friston 2010).

Another important modern-day motivation is to overcome a major constraint in training the current deep learning-based models. Deep learning models require a very large number of samples of the environment that covers a broad spectrum of the reality in order to learn and generalize (LeCun et al. 2015). Collecting these training samples are often not feasible. Interacting directly with the real environment can be expensive, time consuming, and even dangerous in some cases. A learned world model allows an agent to shift part of its learning and decision-making into an internal simulation, with the expectation that it would generalize to the real-world scenarios (Ha and Schmidhuber 2018; Hafner et al. 2019).

In modern AI, world models are typically learned from data and often take the form of probabilistic models that map from latent states and actions to predicted future states and observations (Hafner et al. 2019; Schrittwieser et al. 2020). This means it can also suffer from the same constraint of requiring very large amount of training data. It is also possible that parts of the (latent states + actions → future states + observations) pipeline are not differentiable and therefore gradient-based learning is not possible. In this case, world models need to mix in other gradient-free methods such as heuristic search and predicate logic (LeCun, n.d.).

In summary, a world model is not just a static description of the environment; it is a spatial and temporal model that can be rolled forward in time to support prediction, planning, and "what-if" analysis. It allows agents to learn from imagined experience rather than solely depending on real interactions with the environment.

## 2.2 Historical Foundation

The modern idea of a world model is an internal model that AI agents use to dry run scenarios before choosing a course of actions. This is important because real interactions with the environment are often expensive and may have undesirable side effects. However, this idea did not entirely originate from the field of AI. Rather, it emerged gradually across multiple fields that were all trying to solve a similar problem: how can a system behave intelligently beyond relying solely on immediate sensory input? In engineering, the challenge was in operating physical systems under environments that were not fully under control. In psychology and neuroscience, it was in explaining flexible behavior such as generalization and skilled movement. In AI, it was in planning and decision-making beyond reflexive stimulus-response policies. Historically, world model ideas appeared in:

- Control theory and cybernetics: internal models for robust regulation and prediction (Conant and Ashby 1970; Francis and Wonham 1976; Schwenzer et al. 2021).
- Cognitive science and neuroscience: cognitive maps, mental models, and simulation as mechanisms for intelligent behavior (Tolman 1948; Williams 2018; Johnson-Laird 2004).
- AI and decision theory: explicit environment models for planning (symbolic and probabilistic) and, later, for reinforcement learning (Ha and Schmidhuber 2018; Fikes and Nilsson 1971; Sutton 1991; Schmidhuber 1990).

What is striking is that each of them arrived at a similar thesis: to control or understand a system, you need some internal structure that models its regularities well enough to support prediction and counterfactuals.

The root of world models in control theory and cybernetics was neatly summarized in the article title of (Conant and Ashby 1970): "Every good regulator of a system must be a model of that system". This does not mean the regulator has an exact copy of the system. It would not be practical or even useful that way. Rather, it implies that optimal regulation requires modelling of the dynamics of the system. In control theory, Francis and Wonham (Francis and Wonham 1976) stated the Internal Model Principal that "suitably reduplicated model of the dynamic structure of the disturbance and reference signals" was required in the feedback path to the regulator. In modern engineering control, Model Predictive Control (MPC) is a set of advanced control methods used in many engineering fields. It makes use of a process model to predict the future behavior of the controlled system (Schwenzer et al. 2021). These are very close to the modern world model intuition: good control requires an internal predictive model that guides the actions of the agents.

Cognitive maps and mental models in psychology can also be framed as world models. Tolman (Tolman 1948) argued that behavior in rat maze experiments was better explained by internal map-like representations than by simply sequences of stimulus-response. This was historically important because it suggested that animals (rats) learned structured internal representations (a map) of the environment (maze), which helped them make moves to find the food faster (goal). This cognitive map is essentially a world model, which is a reusable representation that supports inference and planning. The cognitive-map theory contributes two important ideas: 1) world models can be latent structures, and 2) these latent structures enable generalization and planning, not merely simple reactions to stimulations.

Another remarkable direct historical statement of the world model concept was Kenneth Craik's work on explanation and thought, where he framed cognition in terms of small-scale models of external reality that the brain used to make predictions and try out possible scenarios without physically acting on them (Williams 2018). Perception and linguistic comprehension yields mental models, while "thinking and reasoning are internal manipulation of mental models" (Johnson-Laird 2004).

Classical AI treated world models as symbolic logic systems, and explicit models were used for planning and acting. For example, Fikes and Nilsson (Fikes and Nilsson 1971) encoded actions as operators with preconditions and effects such that planning was framed as search through a world model of state changes. While AI's earliest successes in planning required a hand-crafted world model, by the late 20th century, researchers were already proposing learned, differentiable internal models. Sutton's Dyna architecture (Sutton 1991) explicitly integrated learning a model from experience, planning with simulated ("hypothetical") experience, and reactive execution. Schmidhuber (Schmidhuber 1990) used supervised recurrent neural networks to model environment dynamics for planning and reinforcement learning. These were early attempts for learned world models and were direct ancestors of modern learned latent dynamics model used for planning. In the deep learning era, world models are compressed latent variables learned from a large number of data samples and then policies are trained using these learned model, as proposed by Ha and Schmidhuber (Ha and Schmidhuber 2018).

## 2.3 Recent Work

Recent advances in world model research have produced a diverse set of systems, many of which are optimized for a particular domain or simulation setting. Despite this diversity, they have a notable commonality: most of them place strong emphasis on video or image generation and on the visual fidelity of generated content (Xing et al. 2025). This shared focus is especially visible across gaming, 3D scene generation, physical control, and general-purpose video generation, even though these systems differ substantially in architecture, objectives, and domain applications.

In the domain of gaming world models, systems such as Project Genie by Google Labs, Muse by Microsoft Research (Kanervisto et al. 2025), and Oasis by Decart ("Oasis," n.d.) use generative models to simulate video game environments from visual and action inputs. These systems are highly domain specific and produce plausible trajectories and up to minutes of continuous gameplay. There are also general-purpose video generation models such as OpenAI Sora and Google DeepMind Veo. These models are designed to generate high-quality video from textual prompts and, in some cases, prior frames. Their outputs can be visually striking, but strictly speaking they are not world models because they do not explicitly support reasoning and planning.

A related line of work concerns 3D scene, such as World Labs. These systems focus on visually compelling 3D scene generation and egocentric navigation. They represent important advances in 3D scene synthesis by pushing the frontier of spatial realism. They are the first to propose 3D as an interface for humans and machines to "communicate about space, collaborate on designs, and create new environments".

In physical world modeling, systems such as Wayve GAIA-2 (Russell et al. 2025) and NVIDIA Cosmos (NVIDIA et al. 2026) target domains including autonomous driving, robotic manipulation, and embodied navigation. These models demonstrate strong capabilities in representing low-level physics and sensorimotor dynamics across varied environmental conditions such as lighting, weather, and geography.

They are typically tightly coupled to their task domains and often rely on specialized sensors, data pipelines, or control architectures.

A conceptually distinct and increasingly debated direction concerns the level at which a world model should represent and predict the future. On one side, LeCun's joint embedding predictive architecture (JEPA) argues for learning abstract predictive representations rather than reconstructing raw observations (LeCun, n.d.). The JEPA family of systems includes V-JEPA 1 and 2 (Bardes et al. 2024; Assran et al. 2025), DINO-WM (Zhou et al. 2025), and PLDM (Sobal et al. 2025). Instead of predicting and generating realistic environment scenes at the pixel level, these models predict future latent embeddings through encoder-based architectures trained with energy-based objectives. This abstraction-first approach is appealing because it avoids the computational and statistical burden of reconstructing every detail of the observable world and may offer a more tractable route toward task-relevant representation learning. On the other side, Xing's PAN line of work raises a different concern: if latent representations are not sufficiently grounded in observable environments and actionable consequences, they may become difficult to validate, identify, or use reliably for long-horizon planning (Xing et al. 2025; Team et al. 2025). PAN therefore adopts a generative latent prediction architecture that combines autoregressive latent dynamics, grounding in large-scale text knowledge, and a video diffusion decoder, linking language-conditioned reasoning about future states with detailed visual simulation (Team et al. 2025). This debate between LeCun's latent abstraction and Xing's grounded simulation is especially relevant to this work. A BWM does not require reconstruction of the business environment, but it also cannot rely on ungrounded latent representations alone. For business decision-making, the relevant grounding must come from business semantics: entities, attributes, relationships, feasible actions, rules and constraints, and observed business outcomes.

# 3 Business World Model (BWM)

## 3.1 Definitions and Problem Context

A BWM serves as an internal simulator that enables intelligent agents to plan, optimize, and execute business initiatives in pursuit of desired outcomes. It can be understood as a world model specialized for business and organizational environments. Specifically, a BWM encodes:

- organization- and market-specific states, including for example customers, products, demand signals, operational capacity, and business constraints;
- business dynamics, including the strategic and operational relationships among key business factors such as pricing, adoption, churn, revenue, capacity, and cost; and
- the feasible action space available to business initiatives.

As illustrated in Figure 2, by integrating these elements, a BWM allows agents to simulate alternative courses of action, evaluate likely outcomes and side effects, and select actions that are aligned with strategic objectives and operational constraints.

Business initiatives, such as product launches, market expansions, and customer churn reduction, require decisions under partial observability, shifting conditions, and strategic interaction. Unlike the physical environment, where agent behavior can be modeled according to stable and universal physical laws that enable reliable trajectory simulation, business environments lack universally valid and predictive "laws of business". Business outcomes depend on heterogeneous and context-sensitive factors:

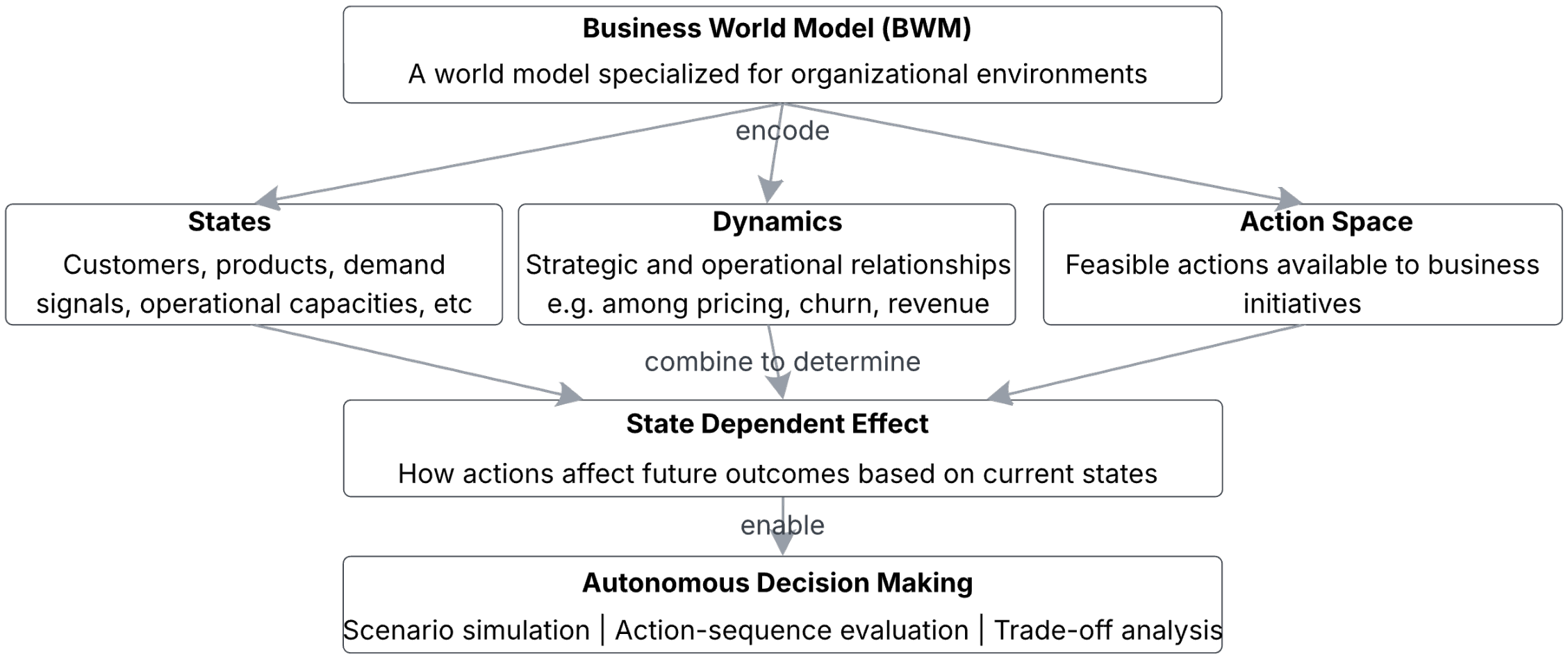

**Figure 2.** Conceptual diagram of a Business World Model (BWM). A BWM is a specialized world model for organizational environments. It operates by encoding current business states, dynamics, and feasible actions to model the state-dependent effects of potential initiatives. By capturing these elements, the architecture provides a foundation for autonomous decision support, enabling intelligent agents to conduct scenario simulations, evaluate action sequences, and perform trade-off analysis under uncertainty.

organizational structure, incentives, regulation, competition, customer behavior, macro conditions, and many more. As a result, business decision support is often fragmented across:

- narrative strategy artifacts (e.g. plans and playbooks),
- descriptive analytics (e.g. dashboards and KPIs), and
- isolated predictive or optimization tools (e.g. forecasting models, pricing models, capacity planning tools).

These components rarely, if ever, compose into a coherent architecture that can be queried, simulated, and reused across business initiatives. This is what a BWM provides.

The key challenge of designing a BWM is in grounding and executability. While large bodies of codified knowledge exist in business management education and practice (e.g., strategy frameworks, operational best practices), they are primarily represented as text and are difficult to operationalize directly. Quantitative tools such as operations research and machine learning models can optimize specific subproblems but typically presume well-defined objectives, variables, and constraints. The central scientific and engineering challenge of architecting a BWM is therefore to:

1) extract, organize, and integrate heterogeneous business knowledge and tools, and
2) ground it first in the context of a business and then in the local context of business initiatives.

BWM provides a systematic framework to this challenge and make the knowledge, data, and tools useful for planning and decision-making by autonomous AI agents.

It is important to distinguish a BWM from large language models (LLMs). LLMs provide reasoning, language understanding, and agentic execution capabilities. However, by themselves, they do not capture the states or dynamics of a business. In BWM, LLMs serve as a reasoning component of the interface, planner and task orchestration.

## 3.2 States, Dynamics and Action Space

We propose a business semantics centric approach to modeling states, dynamics, and action space in BWM, following (Pang 2025). The central idea is that the elements most relevant to business initiatives are those tied to the business's key entities, such as consumer, product, and subscription. A BWM should therefore not attempt to represent every aspect of a business. Instead, it should focus on the aspects of states, dynamics and actions that are useful for scenario simulations in support of autonomous agents in business initiatives. This is in line with the suggestion by LeCun (LeCun, n.d.) that "it would be useful to have a way to bias the system towards representations that contain information relevant to a class of tasks". Linking these elements to business entities is the first step toward grounding them in business context and rendering them queryable and actionable for the planning and execution of business initiatives by autonomous agents.

In order for a goal-directed agent to simulate and plan multi-step trajectory under conditions with uncertainty, it must be able to make probabilistic prediction of the next state based on a given current state and a hypothetical action. Formally:

$$S_{t+1} \sim P(S_{t+1} \mid S_t, a_t)$$

$$a_t \in A(S_t)$$

where $S_t$ is the state at time step $t$; $a_t$ is a hypothetical action taken; and $A(S_t)$ is the space of feasible actions available when the state is $S_t$.

Under a business-semantics-centric formulation, the state of a business $S_t$ is represented by its key entities, the attributes of those entities, and the relationships among them. For example, for a digital subscription company, consumers and their profile attributes represent a central component of the business state. Likewise, the subscriptions associated with those consumers, along with attributes such as monthly rate and tenure of the subscription, constitute an equally important part of that state.

The ability of a business to systematically collect, maintain, and update this semantic information contributes to the quality and overall usefulness of the state representation available to intelligent agents. Given such a state representation, business growth can be understood as the selection of actions that move the business toward more desirable future states $S_{t+n}$. This requires agents to know not only about which actions are feasible $A(S_t)$, but also about how the business state is expected to evolve in response to those actions—that is, about the dynamics of the business system $P(S_{t+1} \mid S_t, a_t)$. In this formulation, feasible actions are actions that a business can take and directly modify the value of one or more attributes of a business entity. For example, changing the price of a product is a feasible action; here, product is the entity, and price is an attribute of product.

The set of feasible actions a business can take is often limited, and therefore not all entity attributes are directly changeable through actions. Indeed, attributes that characterize desirable business outcomes are often not directly controllable. If they were, growing a business would be a trivial optimization problem. For this reason, a BWM must maintain an explicit action space $A(S_t)$ that agents can use to simulate alternatives and support decision-making. The central challenge is then to understand how feasible actions applied to controllable attributes can indirectly influence non-controllable outcome attributes. This is where system dynamics become essential—it describes how entities interact and how states evolve under uncertainty. A BWM must capture how entity attributes are interrelated, how changes propagate across entities and relationships, and ultimately how controllable interventions can move the business toward desired future states.

In contemporary data-driven businesses, dynamics are often approximated through multiple multivariate machine learning models that estimate dynamics among selected entity attributes. Trained on historical data from internal operational systems and external data providers, such models capture recurring patterns in business behavior while accounting for the uncertainty inherent in business environments. They may include narrow scoped models that, for example, predict customer churn probability using pre-selected features such as consumer demographics and online activities. The dynamics may also include broader scoped models such as Graph Neural Networks (GNNs) (Corso et al. 2024) that cover all or subsets of the business entities as graph nodes and their relationships as edges and then predict the future states of selected edges and node attributes. Moreover, business dynamics are not limited to learned models. Depending on the industry and individual businesses, there may be business rules or regulations that are deterministic. These rules can be, for example, coded in logic programs or in structured texts.

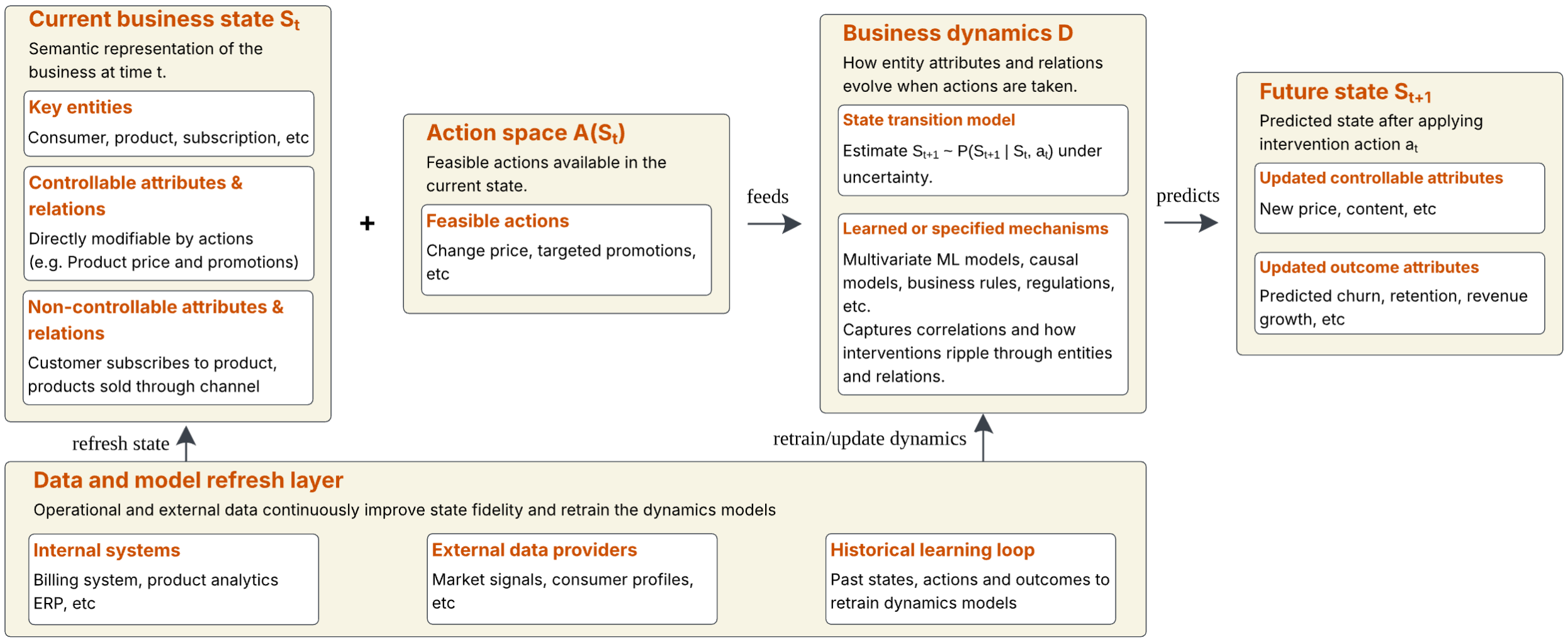

**Figure 3.** States, dynamics, and action space in a Business World Model (BWM). This diagram illustrates the three core components of a BWM. Current state captures the current values of entity attributes, partitioned into controllable attributes that can be directly modified by feasible actions (e.g., product price or discount) and non-controllable attributes that cannot be directly changed (e.g., customer churn risk). Dynamics, typically approximated through machine learning models trained on historical data, or coded business rules, describe how entities interact and how the business state evolves under uncertainty. Together, current state, actions and dynamics determine the future state, such as desirable outcome attributes such as revenue and customer retention. The central challenge: feasible actions can only indirectly influence desirable outcomes through the system's dynamics by acting on controllable attributes.

Figure 3 shows the three core components of a BWM: business states, dynamics and action space.

## 3.3 Incremental Development of BWM Capabilities

An important property of the proposed BWM architecture is that it provides a systematic framework for incrementally expanding its capabilities over time. At an early stage, a business may collect data and build predictive models for only a limited set of entities. As the business grows, additional data sources, entities, and predictive models can be incorporated into the BWM, increasing both the breadth and depth of its planning capabilities. In this way, the BWM can evolve from supporting narrow, localized decisions to enabling more comprehensive reasoning across business functions. This incremental path makes it essential for businesses to adopt a business-semantics-centric data system (Pang 2025) early in their development, so that current and future data collection, modeling, and planning capabilities can be built on a coherent semantic foundation.

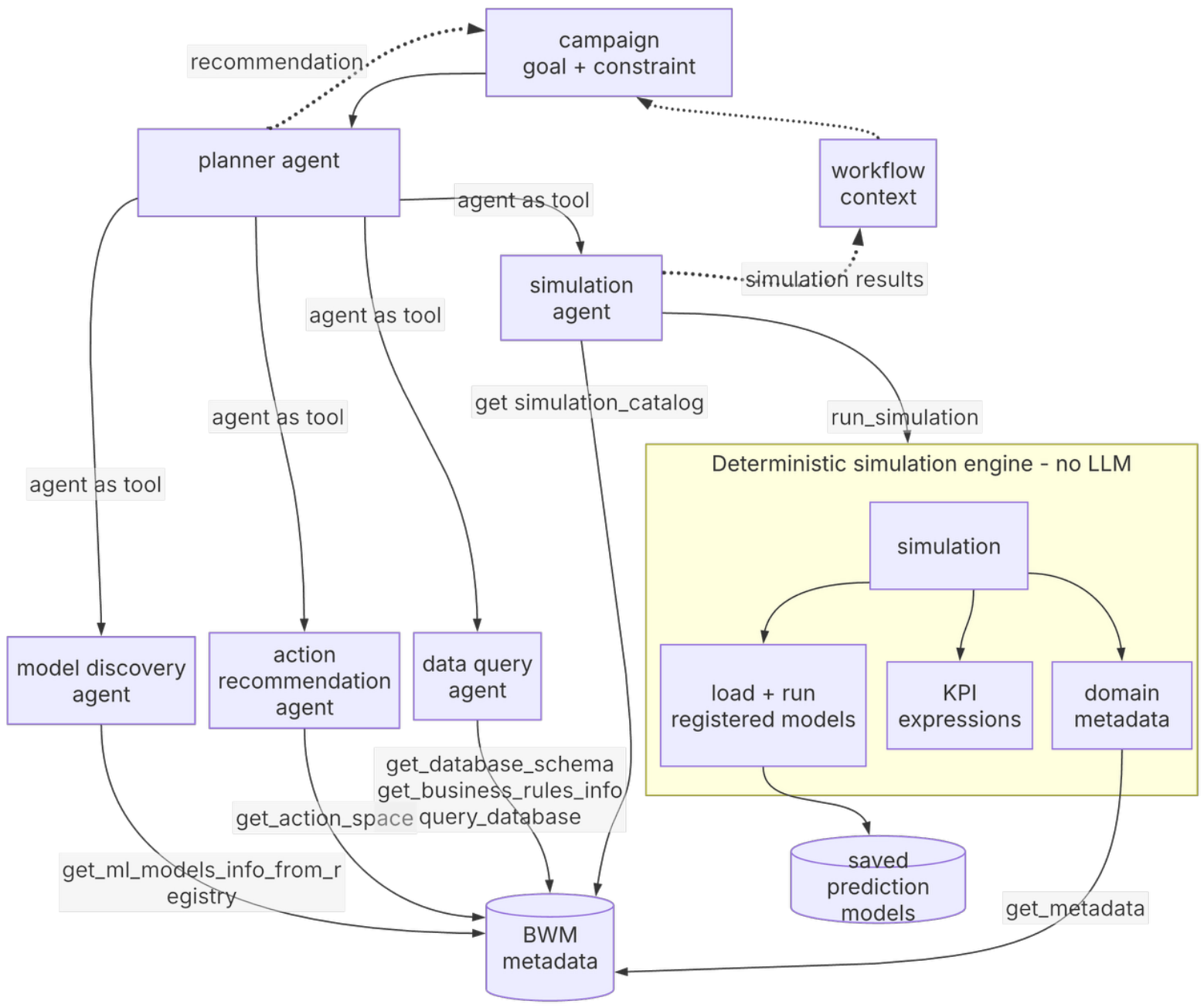

**Figure 4.** BWM system architecture. A planner agent turns a natural-language business goal into a structured recommendation by calling four specialist agents as tools: model discovery, action recommendation, data query and simulation. All domain knowledge comes from the BWM metadata database. The numbers are computed by a deterministic simulation engine that uses no LLM, and its exact results are kept in the shared workflow context (dashed arrows) so the agents' summaries can be checked.

# 4 Reference Design & Case Study

Section 3 describes the conceptual architecture of BWM. This section turns that concept into a concrete design and a working implementation and then validates it with a case study. The full source code with documentation is available on GitHub at: https://github.com/cecilpang/bwm-paper

## 4.1 System Design & Implementation

The system takes a business goal written in natural language and turns it into a structured, verifiable recommendation for action. The design separates reasoning from algorithmic calculation. A set of LLM agents interprets the goal, chooses what to analyze and explains the result. Every calculation and number comes from a deterministic simulation engine. As shown in Figure 4, a planner agent acts as the controller and calls four specialist agents as tools. The model-discovery agent finds the registered prediction models relevant to the goal. The action agent picks the feasible interventions and their allowed levels. The data-query agent answers questions about the data and reports business rules, using single read-only SQL SELECT statements. The simulation agent turns the goal into a formal specification (segment filters, an objective KPI, a target and an optional revenue cap) and calls the simulation engine to run it. The planner returns a typed recommendation that covers the goal interpretation, the baseline, a table of scenarios, a feasibility verdict, the recommended action, trade-offs and caveats. The exact simulation output is also stored in the shared run context and returned with the recommendation, so the LLM's summary can be checked against it. When a tool fails, the agent sees the real error message, which lets it correct its inputs and retry instead of failing silently.

**Algorithm 1.** Deterministic simulation engine

**Input:** specification $S = (F, k^*, \text{scope}, \tau, R_{max})$: segment filters F, objective KPI k*, its scope (segment or population), target $\tau$, optional revenue cap $R_{max}$; BWM metadata M
**Output:** baseline KPIs B, scenario table T with a feasibility verdict per scenario, overall verdict and recommended scenario c*

1: $D \leftarrow$ join the entity tables in M on the unit identifier
2: **for each** prediction model m in M **do**
3:    add column $\hat{y}_m \leftarrow$ transform(score(m, D)) to D
4: $s \leftarrow$ rows of D that satisfy the segment filters F (filters may use the predictions $\hat{y}_m$)
5: $E \leftarrow s$ **if** scope = segment, **else** all rows of D
6: $B \leftarrow$ Evaluate(D), where Evaluate computes every KPI on the rows E, except the revenue-category KPI r, which it computes on all rows
7: $C \leftarrow$ non-empty combinations of action levels (at most one level $\ell \in L(a)$ of each feasible action a)
8: $T \leftarrow \emptyset$
9: **for each** scenario $c \in C$ **do**
10:    $D' \leftarrow D$
11:    **for each** $(a, \ell)$ in c **do** $D'[s, \text{attr}(a)] \leftarrow$ op(a) applied to $D'[s, \text{attr}(a)]$ with $\ell$, where op(a) $\in$ {scale, add, set}
12:    **for each** model m with an input attribute changed by c **do**
13:       re-score m on the rows s of $D'$ (as in line 3)
14:    $K \leftarrow$ Evaluate($D'$)
15:    met $\leftarrow K[k^*] - B[k^*]$ satisfies the target $\tau$
16:    ok $\leftarrow R_{max}$ is undefined **or** $B[r] - K[r] \leq R_{max}$
17:    append (c, K, met, ok) to T
18: $T_{feas} \leftarrow \{t \in T : \text{t.met}$ **and** $\text{t.ok}\}$
19: **if** $T_{feas} \neq \emptyset$ **then** verdict $\leftarrow$ feasible; $c^* \leftarrow$ the scenario in $T_{feas}$ with the smallest revenue loss $B[r] - K[r]$
20: **else** verdict $\leftarrow$ infeasible; $c^* \leftarrow$ none
21: **return** B, T, verdict, c*

Algorithm 1 describes the deterministic simulation engine. Its inputs are the BWM metadata M and a specification S that states the segment of interest (the filters F), the objective KPI k* and whether it is measured over the segment or the whole population, the target $\tau$, and an optional cap $R_{max}$ on how much the revenue KPI may decline. The engine first builds the baseline (lines 1–6): it joins the entity tables into a single table D with one row per unit entity, scores every registered prediction model, selects the segment s with the filters (which may refer to the model predictions), and evaluates the KPIs to obtain the baseline B. It then enumerates the scenarios C, each a combination of action levels (line 7), and simulates them one at a time (lines 9–17). For each scenario, the action levels are applied to the segment rows only, the models whose inputs were changed are re-scored on those rows, and the KPIs are re-evaluated to give K. A scenario is marked as meeting the target (met) if the change in k* relative to B satisfies $\tau$, and as meeting the revenue constraint (ok) if the decline of the revenue KPI r does not exceed $R_{max}$. Finally (lines 18–20), the scenarios for which both conditions hold are feasible, and the engine recommends the feasible scenario with the smallest revenue loss; if no scenario is feasible, the verdict is infeasible. The engine returns the baseline B, the full scenario table T, the verdict, and the recommended scenario c*. Every step is a fixed computation over the data and the registered models, with no sampling and no language model in the loop, so the same inputs always produce the same output.

As shown in the entity-relationship diagram in Figure 5, the BWM metadata describes a business domain declaratively, so the agents and the simulation engine stay free of domain-specific code. The domain is described by entities and their attributes, each with a data type (numeric, categorical, date or identifier). Prediction models are registered with the attribute they populate, the unit entity they score, the location and format of the trained artifact, the prediction kind (probability, value or class), an optional output transform such as prior correction, and a decision threshold. Their predictive features map database attributes to the model's input names and order. Feasible actions name the attribute an intervention changes and how it changes it (scale, add or set), and action levels list the values allowed for each action, for example a scale of 0.85 for a 15% discount. KPIs are named metrics defined as an aggregation (sum, mean, count, min or max) of an expression over entity attributes, including model predictions. Each KPI has a unit and a category (outcome, revenue or volume); the revenue category tells the engine which KPI enforces revenue constraints. Finally, business rules and their links to the entities and attributes they govern capture organizational policies, which are reported to the planner but not enforced. Because every part of the domain is described in these tables, a new model, action, KPI, entity, attribute or an entire new organization can be added by inserting rows, not requiring code change.

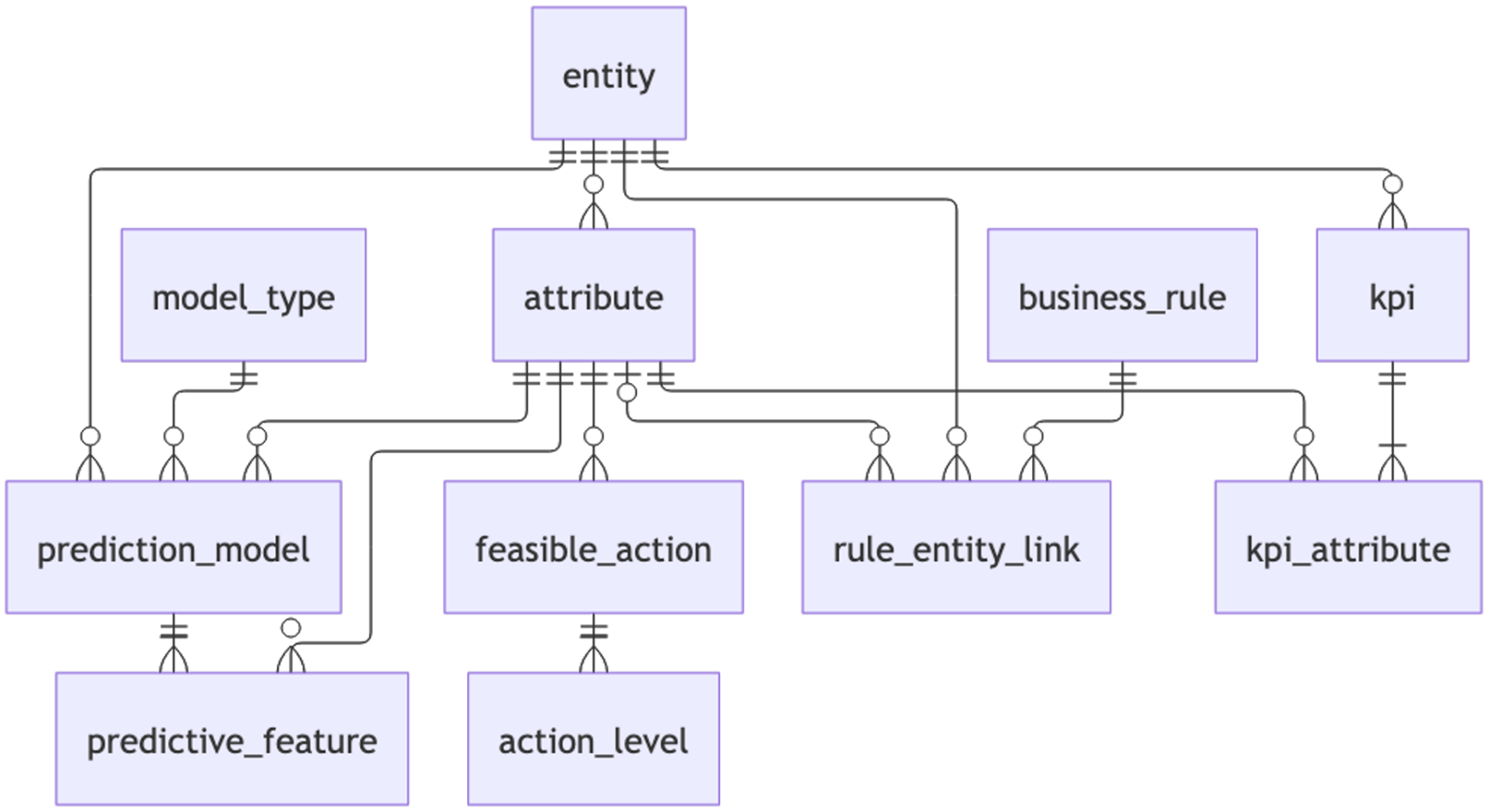

**Figure 5.** The BWM metadata model. Tables describing entities, attributes and their relationships, registered prediction models and their features, feasible actions and their levels, KPIs, and business rules. Every table except model_type is partitioned by organisation.

The reference design shows one simulation engine with configurable parameters. Multiple simulation engines can be plugged into BWM, with each being generic to a broad category of business initiatives in a business enterprise. Equipped with the LLM's reasoning power, the simulation agent selects the appropriate one for the initiative at hand.

## 4.2 Case Study

BWM is validated on a subscription churn case study. The domain has three entities, each with one row per subscription and joined one-to-one on the primary key of subscription. As shown in Figure 6, *subscription* holds the monthly cost and tenure in days. *content_consumption_statistics* holds how often (frequency) and how widely (breadth) the subscriber consumes content. *subscription_predictions* holds the predicted churn probability. The data covers 965,991 subscriptions. A logistic regression churn model is registered as a prediction model that scores subscriptions from frequency, breadth, tenure, and monthly cost. The feasible actions are different levels of price discounts: monthly cost is reduced to 0.95, 0.90, 0.85, 0.80 or 0.75, corresponding to discount range of 5% to 25%. Four KPIs are defined: the mean predicted churn probability and the predicted churn rate (the share of subscriptions above the threshold) as outcome KPIs, the expected monthly revenue as the revenue KPI, and the subscription count as a volume KPI. The system is given the goal:

*"Reduce the churn rate of high frequency subscriptions (top 10%) by at least 0.1 percentage point. Total monthly revenue must not decline by more than $20,000."*

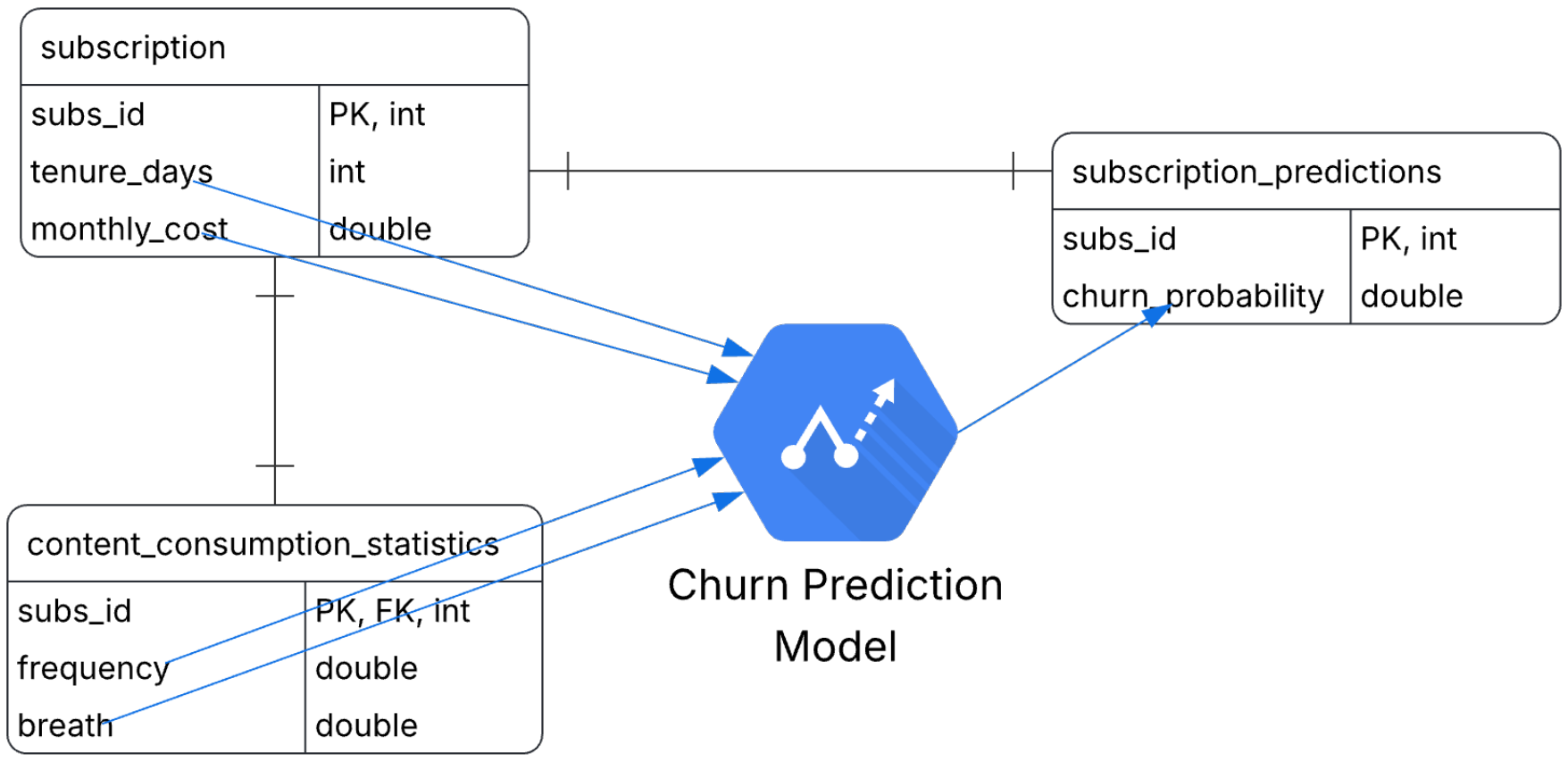

**Figure 6.** E-R diagram of the subscription churn case study. Three entities, each with one row per subscription, are joined one-to-one on the subscription identifier: subscription (monthly cost, tenure), content consumption statistics (frequency, breadth) and subscription predictions (churn probability). The registered churn prediction model takes frequency, breadth, tenure and monthly cost as inputs and predicts the churn probability.

This requires it to find the relevant models, actions and KPIs, define the segment, simulate each discount level, and check each scenario against both the churn target and the revenue constraint.

The case study is added to BWM by only inserting rows into the metadata tables; no tailored code is written for it, including the simulation engine. The three case study tables are registered as *entity* rows, and their nine columns as *attribute* rows. The churn model is registered as a *prediction_model*. Four *predictive_feature* rows map frequency, breadth, tenure days and monthly cost to the model's input columns. The price discount is a *feasible_action* row of type scale, with five *action_level* rows for the values 0.95 to 0.75. The four KPIs are *kpi* rows. The expected monthly revenue KPI is placed in the revenue category, which makes it the KPI that revenue constraints apply to. Because churn probability is defined as a model output rather than plain data, the engine recomputes it whenever an action changes one of the model's inputs. This is how a discount on monthly cost propagates to the churn and revenue KPIs.

The system correctly recommends a 10% discount on monthly cost. It also shows that a discount of 15% would also meet the goal, but with a bigger loss in revenue that is still within the required limit of $20,000.

## 5 Conclusion

This paper introduces the novel concept and architecture of a BWM, provide a reference design and implementation, and a case study to validate the usage of BWM. (Xing et al. 2025) characterizes the development of world models in terms of five key aspects: data, representation, architecture, objective, and usage. This work shows that the same aspects are applicable to the development of BWM. Although many individual components of the proposed architecture are not by themselves new, the contribution lies in integrating them into a coherent architecture that enables intelligent agents to reason about business states, utilize machine learning models to simulate alternative courses of action, and ultimately make autonomous decisions. Our contribution is architectural and systemic.

## References

Assran, Mido, Adrien Bardes, David Fan, et al. 2025. "V-JEPA 2: Self-Supervised Video Models Enable Understanding, Prediction and Planning." Version 1. Preprint, arXiv. https://doi.org/10.48550/ARXIV.2506.09985.

Bardes, Adrien, Quentin Garrido, Jean Ponce, et al. 2024. "Revisiting Feature Prediction for Learning Visual Representations from Video." *Transactions on Machine Learning Research*, May 2. https://openreview.net/forum?id=QaCCuDfBk2.

Conant, Roger C., and W. Ross Ashby. 1970. "Every Good Regulator of a System Must Be a Model of That System." *International Journal of Systems Science* 1 (2): 89–97. https://doi.org/10.1080/00207727008920220.

Corso, Gabriele, Hannes Stark, Stefanie Jegelka, Tommi Jaakkola, and Regina Barzilay. 2024. "Graph Neural Networks." *Nature Reviews Methods Primers* 4 (1): 17. https://doi.org/10.1038/s43586-024-00294-7.

Fikes, Richard E., and Nils J. Nilsson. 1971. "Strips: A New Approach to the Application of Theorem Proving to Problem Solving." *Artificial Intelligence* 2 (3): 189–208. https://doi.org/10.1016/0004-3702(71)90010-5.

Francis, B. A., and W. M. Wonham. 1976. "The Internal Model Principle of Control Theory." *Automatica* 12 (5): 457–65. https://doi.org/10.1016/0005-1098(76)90006-6.

Friston, Karl. 2010. "The Free-Energy Principle: A Unified Brain Theory?" *Nature Reviews Neuroscience* 11 (2): 127–38. https://doi.org/10.1038/nrn2787.

Ha, David, and Jürgen Schmidhuber. 2018. "Recurrent World Models Facilitate Policy Evolution." *Advances in Neural Information Processing Systems* 31. https://proceedings.neurips.cc/paper/2018/hash/2de5d16682c3c35007e4e92982f1a2ba-Abstract.html.

Hafner, Danijar, Timothy Lillicrap, Jimmy Ba, and Mohammad Norouzi. 2019. "Dream to Control: Learning Behaviors by Latent Imagination." Paper presented at International Conference on Learning Representations. September 25. https://openreview.net/forum?id=S1lOTC4tDS.

Hambardzumyan, Sasun, Abhinav Tuli, Levon Ghukasyan, et al. 2022. "Deep Lake: A Lakehouse for Deep Learning." arXiv:2209.10785. Preprint, arXiv, December 13. https://doi.org/10.48550/arXiv.2209.10785.

Johnson-Laird, P. N. 2004. "The History of Mental Models." In *Psychology of Reasoning*. Psychology Press.

Kanervisto, Anssi, Dave Bignell, Linda Yilin Wen, et al. 2025. "World and Human Action Models towards Gameplay Ideation." *Nature* 638 (8051): 656–63. https://doi.org/10.1038/s41586-025-08600-3.

LeCun, Yann. n.d. A Path Towards Autonomous Machine Intelligence Version 0.9.2, 2022-06-27.

NVIDIA, Anis Ahmad, Jon Allen, et al. 2026. "Interactive World Simulation with Cosmos for Physical AI." arXiv:2611.06627. Preprint, arXiv, November 4. https://doi.org/10.48550/arXiv.2611.06627.

"Oasis: A Universe in a Transformer." n.d. Accessed May 24, 2026. https://openreview.net/forum?id=ocMWhGLHqu.

Pang, Cecil. 2025. "Toward Data Systems That Are Business Semantic Centric and AI Agents Assisted." *IEEE Access* 13: 113752–62. https://doi.org/10.1109/ACCESS.2025.3583260.

Pang, Cecil, and Hiroki Sayama. 2026a. "Autonomous Business System via Neuro-Symbolic AI." *2026 IEEE International Systems Conference (SysCon)*, April, 1–8. https://doi.org/10.1109/SysCon66367.2026.11503621.

Russell, Lloyd, Anthony Hu, Lorenzo Bertoni, et al. (arXiv.Org). 2025. "GAIA-2: A Controllable Multi-View Generative World Model for Autonomous Driving." March 26. https://arxiv.org/abs/2503.20523v1.

Schmidhuber, J. 1990. “An On-Line Algorithm for Dynamic Reinforcement Learning and Planning in Reactive Environments.” *1990 IJCNN International Joint Conference on Neural Networks*, June, 253–58 vol.2. https://doi.org/10.1109/IJCNN.1990.137723.

Schrittwieser, Julian, Ioannis Antonoglou, Thomas Hubert, et al. 2020. “Mastering Atari, Go, Chess and Shogi by Planning with a Learned Model.” *Nature* 588 (7839): 604–9. https://doi.org/10.1038/s41586-020-03051-4.

Schwenzer, Max, Muzaffer Ay, Thomas Bergs, and Dirk Abel. 2021. “Review on Model Predictive Control: An Engineering Perspective.” *The International Journal of Advanced Manufacturing Technology* 117 (5): 1327–49. https://doi.org/10.1007/s00170-021-07682-3.

Sobal, Vlad, Wancong Zhang, Kyunghyun Cho, Randall Balestriero, Tim G. J. Rudner, and Yann LeCun. 2025. “Learning from Reward-Free Offline Data: A Case for Planning with Latent Dynamics Models.” Paper presented at The Thirty-ninth Annual Conference on Neural Information Processing Systems. October 29. https://openreview.net/forum?id=hTZ0SJCGQX.

Sutton, Richard S. 1991. “Dyna, an Integrated Architecture for Learning, Planning, and Reacting.” *SIGART Bull.* 2 (4): 160–63. https://doi.org/10.1145/122344.122377.

Team, P. A. N., Jiannan Xiang, Yi Gu, et al. 2025. “PAN: A World Model for General, Interactable, and Long-Horizon World Simulation.” arXiv:2511.09057. Preprint, arXiv, November 15. https://doi.org/10.48550/arXiv.2511.09057.

Tolman, Edward C. 1948. “Cognitive Maps in Rats and Men.” *Psychological Review* (US) 55 (4): 189–208. https://doi.org/10.1037/h0061626.

Williams, Daniel. 2018. “Predictive Minds and Small-Scale Models: Kenneth Craik’s Contribution to Cognitive Science.” *Philosophical Explorations* 21 (2): 245–63. https://doi.org/10.1080/13869795.2018.1477982.

Xing, Eric, Mingkai Deng, Jinyu Hou, and Zhiting Hu. 2025. “Critiques of World Models.” arXiv:2507.05169. Preprint, arXiv, July 27. https://doi.org/10.48550/arXiv.2507.05169.

Zhou, Gaoyue, Hengkai Pan, Yann LeCun, and Lerrel Pinto. 2025. “DINO-WM: World Models on Pre-Trained Visual Features Enable Zero-Shot Planning.” Paper presented at Forty-second International Conference on Machine Learning. June 18. https://openreview.net/forum?id=D5RNACOZEI.